\documentclass{article}
\usepackage{amssymb}
\usepackage{booktabs}
\usepackage{amsfonts}
\usepackage{amsmath}
\usepackage{amssymb}
\usepackage{amsthm}
\usepackage{nicefrac}
\usepackage{xcolor}
\usepackage{mathtools}
\usepackage{bm}
\usepackage{multirow}

\newtheorem{theorem}{Theorem}

\theoremstyle{definition}

\theoremstyle{remark}
\newtheorem{remark}[theorem]{Remark}

\newcommand{\R}{\mathbb{R}}
\newcommand{\E}{\mathbb{E}}
\newcommand{\KL}{\mathrm{KL}}
\newcommand{\TV}{\mathrm{TV}}
\newcommand{\FI}{\mathcal{I}}

\newcommand{\PP}{\mathcal{P}}

\newcommand{\tr}{\mathrm{tr}}

 \usepackage[preprint]{neurips_2026}

\usepackage[utf8]{inputenc} 
\usepackage[T1]{fontenc}    
\usepackage{hyperref}       
\usepackage{url}            
\usepackage{booktabs}       
\usepackage{amsfonts}       
\usepackage{nicefrac}       
\usepackage{microtype}      
\usepackage{xcolor}         

\title{Energy Generative
Modeling: A Lyapunov-based Energy Matching Perspective}

\author{%
  Yixuan Wang\thanks{Corresponding Author.} \quad
  Wenqian Xue \quad
  Warren E. Dixon \\
  Department of Mechanical and Aerospace Engineering \\
  University of Florida \\
  \texttt{\{wang.yixuan, w.xue, wdixon\}@ufl.edu}
}

\begin{document}
\maketitle
\begin{abstract}
    Generative models based on static scalar energy functions represent an emerging paradigm in which a single time independent potential drives sample generation through its gradient field, eliminating the need for time conditioning entirely.
    We unify the training and sampling phases of this paradigm, conventionally treated as separate procedures, within a single framework: density transport on the Wasserstein space, cast as a nonlinear control problem in which the Kullback Leibler (KL) divergence serves as a Lyapunov function.
    Training and sampling are then two instances of this same master dynamics, differing only in initial condition.
    Within this autonomous framework we develop two analytic results.
    First, since the Lyapunov certificate is asymptotic, we derive a finite step stopping criterion for Langevin sampling and prove that no Lyapunov certificate exists for the deterministic gradient flow on the same energy landscape.
    Second, the reformulation brings the toolkit of nonlinear control theory to bear on static scalar energy generative modeling, that is, we show that additive composition of trained scalar energies retains an explicit Gibbs invariant measure and inherits the closed-loop Lyapunov certificate.
    Beyond these immediate results, this reformulation bridges static scalar energy generative models with the full toolkit of nonlinear control theory, opening the door to barrier functions for constrained generation and contraction metrics for accelerated sampling.
    Experiments on synthetic distributions validate the theoretical predictions.
\end{abstract}

\section{Introduction}
\label{sec:intro}

The primary task for generative models is to construct a mechanism that produces samples from an optimal distribution $\rho^*$, which is an approximation of the underlying data distribution, given only a finite training set $\{x_i\}_{i=1}^N \subset \R^d$.
Diffusion models~\cite{sohl2015deep,ho2020ddpm,song2019generative,song2021sde} and flow matching~\cite{lipman2023flow,liu2023rectified,albergo2023stochastic,tong2024improving} have emerged as the dominant approach, learning time dependent vector fields or score functions that transport a simple noise distribution to the data distribution along a continuous probability path.
%
%
A recent paradigm, Energy Matching \cite{balcerak2025energy}, has emerged that removes time conditioning entirely.
Instead of learning time varying dynamics, one learns a single scalar energy function $U_\theta \colon \R^d \to \R$, defines a Gibbs distribution $\rho_{\theta}(x) = \exp(-U_\theta(x))/Z_\theta$ as a surrogate for $\rho_{\mathrm{data}}$, and draws samples by running dynamics on this static energy landscape.
%
%

The static energy paradigm possesses a structural property that the time conditioned paradigm does not: it produces an autonomous (time independent) dynamical system.
Concretely, samples are drawn by simulating the stochastic differential equation $dx_t = -\nabla U(x_t) dt + \sqrt{2} dW_t$, whose distributional counterpart is the Fokker Planck equation $\partial_t \rho_t = \nabla \cdot (\rho_t \nabla U_\theta) + \Delta\rho_t$, both governed by the time invariant drift $\nabla U_\theta$.
Autonomous systems are precisely the setting where the classical toolkit of nonlinear control theory applies most naturally: Lyapunov stability and control barrier functions~\cite{khalil2002nonlinear,sontag1998mathematical,ames2017control}.

The classical analytical perspective on the Langevin sampler can be viewed as treating $\KL(\rho_t \| \rho_\theta)$ as a Lyapunov function on $\PP_2(\R^d)$ \cite{bakry2014analysis}.
This is an analytical guarantee for a dynamics inherited from physics.
However, it does not address why the Langevin stochastic differential equation (SDE), rather than its deterministic counterpart $\dot x_t = -\nabla U_\theta(x_t)$ or any other autonomous flow on the same energy landscape, is the correct distributional generator in the first place.
The generative modeling objective is fundamentally distributional: the task is not to move one particle to one mode but to steer a prior distribution $\rho_0$ toward $\rho^*$ such that minimizes a metric such as total variation or KL divergence.

The present work reformulates the static energy generative problem as a nonlinear control problem of probabilistic density transportation on the space of $\PP_2(\R^d)$.
In this reformulation, the state variable is the probability density $\rho_t$ itself, the open-loop dynamics are governed by the continuity equation $\partial_t \rho_t = -\nabla \cdot (\rho_t u_t)$ with the velocity field $u_t$ serving as the control input, and the KL divergence $\KL(\rho_t \| \rho^*)$ serves as the Lyapunov function.
The Lyapunov design requirement that $\frac{d}{dt}\KL(\rho_t \| \rho^*) \leq 0$ along all trajectories determines a unique optimal velocity field, and the resulting closed-loop density dynamics is the Fokker-Planck equation.

Our contributions are as follows.
\begin{enumerate}
\item We develop a control-theoretic reformulation framework in which training, sampling, and smoothing are unified as three instances of density transport on $P_2(\R^d)$, each governed by a continuity equation.

\item We provide a finite time stopping analysis for deterministic sampling and Langevin sampling.
The deterministic flow is analyzed as an early stopped sampling surrogate, despite its asymptotic tendency toward mode collapse, while Langevin dynamics is treated as the intrinsic stochastic sampler.
Our analysis identifies when deterministic transport can be terminated in a meaningful distributional regime and uses a Lyapunov based argument to characterize the performance of Langevin sampling.

\item We bring the toolkit of nonlinear control theory to bear on static scalar energy generative modeling.
We show that additive composition of trained scalar energies retains an explicit Gibbs invariant measure and inherits the closed-loop Lyapunov certificate, which enables us the tool of control barrier function for safety guaranteed generation.

\end{enumerate}

\paragraph{Notation.}
Throughout, $\nabla$ denotes the gradient operator $(\partial_{x_1}, \ldots, \partial_{x_d})^\top$ and $\nabla \cdot$ denotes the divergence operator acting on vector fields: $\nabla \cdot F = \sum_{i=1}^d \partial_{x_i} F_i$.
The Laplacian is $\Delta = \nabla \cdot \nabla = \sum_{i=1}^d \partial_{x_i}^2$.
The space $\PP_2(\R^d)$ consists of Borel probability measures on $\R^d$ with finite second moment.
For $\rho, \nu \in \PP_2(\R^d)$ with $\rho$ absolutely continuous with respect to $\nu$, the KL divergence is $\KL(\rho \| \nu) = \int_{\R^d} \rho \log(\rho/\nu)\,dx$, the total variation distance is $\TV(\rho, \nu) = \frac{1}{2}\int_{\R^d}|\rho - \nu|\,dx$, and the relative Fisher information is $\FI(\rho \| \nu) = \int_{\R^d} \rho \,\|\nabla \log(\rho/\nu)\|^2\,dx$. {For $A \in \mathbb{R}^{n \times n}$, $A \succeq 0$ denotes a semi-positive definite matrix $A$.}
$K_h$ is a symmetric positive definite kernel used in the kernal density estimation (KDE) with bandwidth $h>0$, for example the Gaussian kernel $K_h(x) = (2\pi h^2)^{-d/2} exp(-\|x\|^2 / (2h^2))$.
$x$ denotes the data input and $\theta$ denotes the parameters for the neural network.

\section{Control-Theoretic Reformulation}
\label{sec:control}
In this section, we develop the control-theoretic reformulation.
Take the Wasserstein space $\PP_2(\R^d)$ as the state space and the probability density $\rho \in \PP_2(\R^d)$ as the state variable. The open-loop dynamics of the probability density are governed by the continuity equation
\begin{equation}\label{eq:continuity}
  \partial_t \rho \;=\; -\nabla \cdot (\rho\, u),
\end{equation}
in which the velocity field $u \colon \R^d \to \R^d$ enters as the control input.
The design objective is to select a controller $u$ such that the resulting closed-loop dynamics converges a prior distribution $\rho_0$ to a target distribution.
The prior distribution and target distribution are different between the training phase and sampling phase.
In training phase, a neural network learns an static scalar energy field $U_\theta^*$, such that the corresponding Gibbs density $\rho_\theta^*(x) = \exp(-U_\theta^*(x))/Z_\theta^*$ is an optimal approximation of $\rho^*$ as $t \to \infty$.
In sampling phase, a well trained energy field $U_{\theta}^*$ is given.
The density transportation is between the prior noise distribution $\rho_0$ and $\rho_\theta^*$.
Since both training and sampling share the same dynamical system, we take the training phase as an example in the later derivations. 

To accomplish this objective through Lyapunov's direct method, we choose the KL divergence as the Lyapunov candidate
\begin{equation}\label{eq:lyapunov_candidate}
  V(\rho_\theta) = \KL(\rho_\theta \| \rho^*) = \int_{\R^d} \rho_\theta(x) \log \frac{\rho_\theta(x)}{\rho^*} dx.
\end{equation}
%
Then, following control theory, the Lie derivative \cite{Lie} of $V$ along the open-loop dynamics in $\eqref{eq:continuity}$ is computed by (see Appendix \ref{app:lyapunov_deriv})
\begin{equation}\label{eq:lie_derivative}
  \mathcal{L}_u V(\rho_\theta) = 
  \int_{\R^d} \rho_\theta(x)\, u(x) \cdot \bigl(\nabla \log \rho_\theta(x) + \nabla U^*\bigr)\,dx,
\end{equation}
%
%
where $U^*$ corresponds to $\rho^*$ in the Gibbs density. The same expression appears in the Wasserstein gradient flow theory \cite{jordan1998variational,villani2009optimal}, here reached through the control-theoretic route of computing the directional derivative of a Lyapunov candidate along the controlled flow.
To enable convergence, the Lyapunov condition $\mathcal{L}_u V(\rho_\theta) < 0$ for $\rho_\theta \neq \rho^*$ is required to be met.
%
By selecting the optimal controller $u$ as
\begin{equation}\label{eq:optimal_control}
  u^*(x,\rho_\theta) = -\nabla U^*(x) - \nabla \log \rho_\theta(x),
\end{equation}
the Lyapunov derivative collapses to the negative relative Fisher information,
\begin{equation}\label{eq:fisher_decay}
  \mathcal{L}_{u^*} V(\rho_\theta) 
  = -\int_{\R^d} \rho_\theta(x)\,\bigl\|\nabla \log \rho_\theta(x) + \nabla U^*(x)\bigr\|^2 dx
  = -\FI(\rho_\theta \|\rho^*) \leq 0,
\end{equation}
where the last equal holds if and only if $\rho_\theta = \rho^*$.
Then, from the Lyapunov stability principles \cite{khalil2002nonlinear}, the controlled trajectory satisfies $\rho_\theta \to \rho^*$ as $t \to \infty$.
The updated neural network parameter $\theta$ will update the controller at each update cycle, as discussed later.

The closed-loop density dynamics, obtained by substituting \eqref{eq:optimal_control} into \eqref{eq:continuity} and using $\rho \nabla \log \rho = \nabla \rho$, is the Fokker Planck equation
\begin{equation}\label{eq:fokker_planck}
  \partial_t \rho = \nabla \cdot (\rho\,\nabla U^*) + \Delta \rho,
\end{equation}
whose particle level realization is the Langevin SDE~\cite{risken1996fokker}
\begin{equation}\label{eq:langevin}
  dx_t = -\nabla U^*\,dt + \sqrt{2}\,dW_t,
\end{equation}
where $W_t$ is the standard Brownian noise.

On the sampling side, a Langevin SDE is recovered by the steepest descent controller selected by the Lyapunov design.
The optimal control law $u^*$ admits a decomposition into a feedforward term $-\nabla U_\theta^*$, encoding the score of the target measure, and a state-dependent feedback term $-\nabla \log \rho_\theta$ that the closed-loop system in \eqref{eq:langevin} realizes through the Brownian increment $\sqrt{2}\,dW_t$.
Removing Brownian noise produces deterministic gradient descent on the learned energy field $U_\theta^*$, which, as we rigorously establish in Section~\ref{sec:deterministic}, cannot
serve as a distributional sampler.

%

%
%
On the training side, plugging $\rho_\theta$ into \eqref{eq:optimal_control}, identifying the velocity field $u^*(x, \rho_\theta) = -\nabla U^*(x) + s_{\rho_\theta} (x)$, where $s_{\rho_\theta} \triangleq \nabla \log \rho_\theta$ is the score of the model density from $U_\theta$, is equivalent to fit the energy $U_\theta$.
To enable the convergence $\rho_\theta \to \rho^*$, fitting $U_\theta$ corresponds to two matching contributions. One is the implicit score matching that minimizes $\E_{\rho^*}\bigl[\tfrac{1}{2}\|\nabla U_\theta\|^2 - \Delta U_\theta\bigr]$ and recovers the score by integration by parts identity \cite{hyvarinen2005estimation}.
The other one is denoising score matching that minimizes $\E_{\rho^*, q_\sigma}\bigl[\|\nabla U_\theta(\tilde x) - \nabla \log q_\sigma(\tilde x \mid x)\|^2\bigr]$ under a perturbation kernel $q_\sigma$ \cite{vincent2011connection}.
Both losses, optimized over a sufficiently expressive scalar parameterization, drive $\rho_\theta \to \rho^*$ in the metric induced by the relative Fisher information.
Therefore, training and sampling are not two separate algorithmic blocks but two specifications of the same Lyapunov controller within a unified framework: the training stage approximates $\rho^*$, while the sampling stage applies the full controller $u^*$ to a noise initial condition.

%
Regarding the form of available data to provide the training target $\rho^*$, the empirical measure as a Dirac representation $\hat\rho_{\mathrm{data}} = \frac{1}{N}\sum_{i=1}^N \delta_{x_i}$ is singular and admits neither a score nor a finite KL divergence to any absolutely continuous density.
Therefore, the effective target of score-based or KL-based training procedures is a smoothed density $\rho^* = \hat\rho_{\mathrm{data}} \ast K_h$, obtained by convolution with a symmetric strictly positive kernel $K_h$ of bandwidth $h > 0$ \cite{parzen1962estimation}.
The smoothing operation $\hat\rho_{\mathrm{data}} \mapsto \rho^*$ is itself a kernel operation on $\PP_2(\R^d)$ in the limiting sense that convolution with a diffusion semigroup interpolates between the singular and the smoothed measures.
%
%
The smoothed density $\rho^* = \hat\rho_{\mathrm{data}} \ast K_h$ enters the framework as the fixed target of the training transport.
The training process then seeks the optimal training result such that $\rho_{\theta}^* $ is an close approximation of $ \rho^*$, which closes the framework, identifying the Lyapunov target with the smoothed data measure and the closed-loop sampler \eqref{eq:langevin} as a transport whose stationary distribution is exactly $\rho^*$.
Table~\ref{tab:three_components} summarises the transports unified by this construction.

\begin{table}
\small
\centering
\begin{tabular}{ccc}
\toprule
Transport & Initial state & Final state\\
\midrule
Training           & $\rho_{\theta_0}$                     & $\rho^*_{\theta}$\\
Langevin sampling  & $\rho_0$ (Gaussian)                   & $\rho^*_{\theta}$\\
\bottomrule
\end{tabular}
\caption{Two density transports on $\PP_2(\R^d)$ unified by the control-theoretic reformulation.
Both are governed by a continuity equation.
%
}
\label{tab:three_components}
\end{table}

\begin{remark}[Observer interpretation of smoothing]\label{rem:observer}
The map $\{x_i\}_{i=1}^N \mapsto \rho^*$ admits a natural reading in the language of nonlinear estimation.
View the unknown data generating density $p_{\mathrm{data}}$ as the unknown object to be estimated, and the i.i.d. sample $\{x_i\}_{i=1}^N$ as the measurement of this object.
The kernel density estimate $\rho^*$ then plays the role of a nonparametric observer, that is, it produces from the measurement an explicit estimate of $p_{\mathrm{data}}$, and the classical $L^1$ consistency
\begin{align*}
    \mathbb{E}\,\|\rho^* - p_{\mathrm{data}}\|_{L^1(\R^d)} \rightarrow 0 \quad \text{as} \quad N \to \infty,\; h(N) \to 0,\; N h(N)^d \to \infty
\end{align*}
is the convergence of this estimate to its target.
\end{remark}

\section{Sampling Convergence Rates and Finite Step Stopping}
\label{sec:deterministic}

\subsection{Deterministic Sampling Leads to Mode Collapse}
\label{sec:nogo}

Given a learned energy $U_\theta^*$, setting the noise to zero in \eqref{eq:langevin} produces the deterministic gradient flow $\dot x_t = -\nabla U^*_\theta(x_t)$, whose distributional counterpart is the continuity equation
\begin{equation}\label{eq:det_flow}
  \partial_t \rho_t = \nabla \cdot (\rho_t \, \nabla U^*_\theta),
\end{equation}
corresponding to the controller given by $u^*_d(x)= -\nabla U_\theta^* (x)$. This dynamics drives every individual trajectory toward a critical point of $U_\theta^*$, concentrating the distribution onto local minima, a phenomenon known as mode collapse.
%
%
To analyze the failure to preserve the target distribution in this case, 
%
substituting the Gibbs density $\rho_\theta^* = e^{-U_\theta^*}/Z_\theta^*$ into the right hand side of \eqref{eq:det_flow} and using $\nabla \rho_\theta^* = -\rho_\theta^* \nabla U_\theta^*$ gives
\begin{equation}\label{eq:det_residual}
  \nabla \cdot (\rho_\theta^* \nabla U_\theta^*)
  = (-\rho_\theta^* \nabla U_\theta^*) \cdot \nabla U_\theta^* + \rho_\theta^*  \Delta U_\theta^* =
   \rho_\theta^*  h(x),
\end{equation}
where $h(x)\triangleq \Delta U_\theta(x)^* - \|\nabla U_\theta(x)^*\|^2$ is a residual function {(see details in Appendix \ref{app:fixed_point})}.
At any nondegenerate local minimum $x_0$, one has $h(x_0) = \mathrm{tr}(\nabla^2 U_\theta^*(x_0)) > 0$, while under the assumption that $U_\theta^*$ has Lipschitz gradient and confining tail forces {$\|\nabla U_\theta^*(x)\|^2 \to \infty$} as $\|x\| \to \infty$, we have $h(x) \to -\infty$. Therefore, $h$ changes sign and cannot vanish identically.
{It follows that $\partial_t \rho^*_\theta \neq 0$, so the Gibbs density is not a stationary solution of the deterministic flow in} \eqref{eq:det_flow}, and the deterministic flow possesses no equilibrium that coincides with the target.
{We now show a stronger result} extending this local failure to a global obstruction. 

\begin{theorem}[Failure of deterministic Sampling]\label{thm:no_certificate}
Let $U_\theta^* : \R^d \to \R$ have Lipschitz gradient and satisfy $\|\nabla U^*_\theta(x)\| \to \infty$ as $\|x\| \to \infty$. {Consider the deterministic density evolution in \eqref{eq:det_flow}.}
There exists no Lyapunov functional $V : \PP_2(\R^d) \to \R_{\ge 0}$ that simultaneously satisfies (i) $V(\rho) = 0$ if and only if $\rho = \rho_\theta^*$, (ii) $V(\rho) \ge 0$ for all $\rho \in \PP_2(\R^d)$, and (iii) $V(\rho_t)$ nonincreasing along every solution of \eqref{eq:det_flow}.
\end{theorem}

The proof can be seen in Appendix~\ref{app:det_proofs}. 
The interpretation is that {the entropy gradient term $-\nabla \log \rho_t$,} 
which is realized at the particle
level through the Brownian noise $\sqrt{2}\,dW_t$ in \eqref{eq:langevin}, {is an essential component of the optimal transport mechanism} in Section~\ref{sec:control}.
Removing this term, Theorem~\ref{thm:no_certificate} finalizes that no Lyapunov function can exist under the deterministic flow.

\subsection{Convergence Rate of the Langevin Dynamics}
\label{sec:rate}

\begin{figure}
    \centering
    \includegraphics[width=\linewidth]{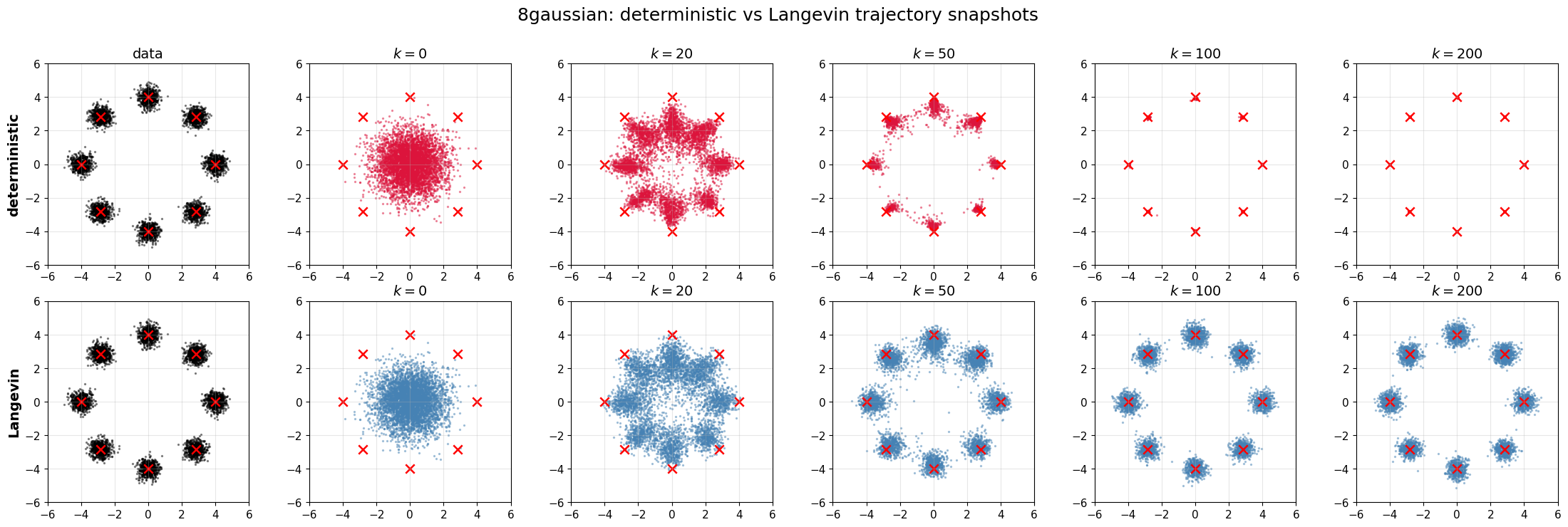}
    \caption{Trajectory snapshots on the eight Gaussian targets with sampling step $K = 200$.
    \textbf{Top row, deterministic flow}. The empirical distribution is closest to $\rho^*$ at $k=50$, and degenerates into a sum of Dirac masses on the eight critical points of $U_\theta^*$ as $k$ increases, in agreement with Theorem~\ref{thm:no_certificate}.
    The deterministic stopping rule in \eqref{eq:det_stop} predicts $k^* = 49$, marked by the proximity of the $k = 50$ panel to the data.
    \textbf{Bottom row, Langevin SDE}. The empirical distribution converges to the Gibbs distribution $\rho_\theta^*$ by $k \approx 100$ and remains visually unchanged at $k = 200$. Additional integration past $\tau_{\rm Lang}^*$ does not degrade the bound.}
    \label{fig:det_vs_lang}
\end{figure}

Given $\rho_\theta^*$ obtained by the trained model, the Lyapunov candidate for sampling is given by
\begin{equation}\label{eq:sampLyap}
    V(\rho_t)=\KL(\rho_t \,\|\, \rho_\theta^*).
\end{equation}
The Lyapunov derivative $\mathcal{L}_u V(\rho_t)  = -\FI(\rho_t \,\|\, \rho_\theta^*)$ established in \eqref{eq:fisher_decay} {provides a qualitative convergence guarantee for $\rho_t \rightarrow \rho_\theta^*$. To obtain the convergence rate, we invoke a functional inequality relating the KL divergence and Fisher information. Specifically,} the energy $U_\theta^*$ is said to satisfy the log Sobolev inequality with a constant $C_{\mathrm{LS}} > 0$ when
\begin{equation}\label{eq:lsi}
  \mathcal{L}_{u^*}V(\rho_t)
  \;\le\;
  \frac{1}{2 C_{\mathrm{LS}}}\,\FI(\rho_t \,\|\, \rho_\theta^*)
  \quad
  \text{for every smooth density } \rho_t \in \PP_2(\R^d).
\end{equation}
{This} condition is guaranteed by the Bakry--\'Emery criterion~\cite{bakry1985diffusions,bakry2014analysis} when
$\nabla^2 U_\theta^*(x) \succeq C_{\mathrm{LS}}\, I_d$ for all $x$. 
Combining \eqref{eq:lsi} with the Lyapunov derivative in \eqref{eq:fisher_decay} on the infinite
dimensional state space $\PP_2(\R^d)$ yields
\begin{equation}\label{eq:exp_convergence}
   V(\rho_t)
  \;\le\;
  e^{-2 C_{\mathrm{LS}} t}\,V(\rho_0),
\end{equation}
which implies exponential convergence of the KL divergence $\KL(\rho_t \,\|\, \rho_\theta^*)$ along the closed-loop dynamics.
The rate in \eqref{eq:exp_convergence} {depends on the log Sobolev constant $C_{\text{LS}}$}, which can degrade significantly in multimodal settings.
In particular, for energy landscapes with barriers of height $\Delta E$, one typically has $C_{\mathrm{LS}} \sim e^{-2 \Delta E}$, leading to exponentially slow convergence \cite{holley1987logarithmic}.
This limitation is inherent to analyses based on \eqref{eq:lsi} and motivates
the empirical surrogate {and deterministic stopping rule introduced in} the next subsection. 

\subsection{Stopping Criteria for Sampling}
\label{sec:stopping}

The convergence rate in \eqref{eq:exp_convergence} prescribes a stopping time of Langevin sampling given the trained model $\rho_\theta^*$.
The time required to achieve a prescribed generation accuracy $V(\rho_t) \le \varepsilon$ can be obtained by solving \eqref{eq:exp_convergence} for $t$
\begin{equation}\label{eq:langevin_stop}
  \tau_{\mathrm{Lang}}^*
  \;=\;
  \frac{1}{2 C_{\mathrm{LS}}}\,\log\!\Bigl(\frac{V(\rho_0)}{\varepsilon}\Bigr),
\end{equation}
at which the bound $\KL(\rho_{\tau_{\mathrm{Lang}}^*} \,\|\, \rho^*_\theta) \le \varepsilon$
holds by substituting $t=\tau_{\mathrm{Lang}}^*$ into \eqref{eq:exp_convergence}.
This result aligns with asymptotic convergence rate analysis. 
Equation \eqref{eq:langevin_stop} is a sufficient sampling time, not a stopping rule in the operational sense: the Lyapunov decay in \eqref{eq:fisher_decay} guarantees that $\KL(\rho_t \,\|\, \rho^*_\theta)$ is decreasing along the closed-loop, so running the Langevin sampler past $\tau_{\mathrm{Lang}}^*$ never degrades the bound.
The role of \eqref{eq:langevin_stop} is to certify that for a given performance bound $\epsilon$, the sufficient sampling time is $t=\tau_{\mathrm{Lang}}^*$.

%

By Theorem~\ref{thm:no_certificate}, no Lyapunov certificate of $\rho^*_\theta$ exists for \eqref{eq:det_flow}, and running the deterministic flow longer eventually drives $\rho_t$ away from the target through mode collapse onto the critical set of $U_\theta^*$.
A finite step analysis is nevertheless possible if one accepts that the goal is no longer to certify convergence to $\rho^*_\theta$, but to identify the latest moment at which the deterministic trajectory still represents a meaningful approximation to $\rho^*_\theta$ before Theorem~\ref{thm:no_certificate} takes effect\footnote{The convergence failure of deterministic sampling in Theorem~\ref{thm:no_certificate} is a distributional statement; at the particle level, the deterministic flow concentrates each trajectory onto a critical point of $U_\theta^*$, which often coincides with a high quality region of the data manifold and yields visually plausible samples in high dimensional practice, although mode coverage is generically lost.}.
Differentiating the Lyapunov function $V(\rho_t)$ in \eqref{eq:sampLyap} along the deterministic evolution $\partial_t \rho_t = \nabla \cdot (\rho_t \nabla U^*_{\theta})$ and applying integration by parts gives {(see Appendix~\ref{app:stein})}
\begin{equation}\label{eq:det_lyap}
  \mathcal{L}_{u^*} V(\rho_t)\;=\;
  \E_{\rho_t}[\Delta U^*_{\theta}] \;-\; \E_{\rho_t}[\|\nabla U^*_{\theta}\|^2]
  \;=\;
  -\,\E_{\rho_t}[\|\nabla U^*_{\theta}\|^2]\,(1 - R_t),
\end{equation}
where the dimensionless ratio
\begin{equation}\label{eq:Rt}
  R_t \;:=\; \frac{\E_{\rho_t}[\Delta U^*_{\theta}]}{\E_{\rho_t}[\|\nabla U^*_{\theta}\|^2]}
\end{equation}
{compares the Laplacian term with the drift-induced dissipation.}
The threshold value $R_t=1$ is the equality case of \eqref{eq:det_lyap} and crossing this value marks the regime change from drift dominated dissipation to Laplacian dominated concentration.
However, the equality $\E_{\rho_t}[\Delta U^*_{\theta}] = \E_{\rho_t}[\|\nabla U^*_{\theta}\|^2]$ does not imply $\rho_t = \rho^*_\theta$. The deterministic flow therefore exhibits two regimes: a drift dominated phase $R_t < 1$ in which the KL divergence decreases, 
and a Laplacian dominated phase $R_t > 1$ in which the KL divergence increases, indicating concentration of mass near local minima. The natural stopping time is the transition point
\begin{equation}\label{eq:det_stop}
  \tau_{\mathrm{det}}^*
  \;=\;
  \inf\{\, \tau \ge 0 \,:\, R_\tau = 1 \,\},
\end{equation}
{at which the KL divergence derivative changes sign and attains its minimum along the trajectory.
Integrating the entropy derivative in \eqref{eq:det_lyap} up to this stopping time yields}
\begin{equation}\label{eq:det_bound}
  \KL(\rho_{\tau_{\mathrm{det}}^*} \,\|\, \rho^*_\theta)
  \;=\;
  \KL(\rho_0 \,\|\, \rho^*_\theta)
  \;-\; \int_0^{\tau_{\mathrm{det}}^*} \E_{\rho_t}[\|\nabla U^*_{\theta}\|^2]\,(1 - R_t)\,dt,
\end{equation}
which represents the maximum decrease of KL divergence achievable under deterministic evolution.

The two stopping times in \eqref{eq:langevin_stop} and \eqref{eq:det_stop} play structurally different roles.
The Langevin time in \eqref{eq:langevin_stop} is a sufficient sampling budget, justified by the Lyapunov certificate of Section~\ref{sec:control}, and running past it is harmless.
While, running past the deterministic time in \eqref{eq:det_stop} strictly degrades the approximation.
Figure~\ref{fig:det_vs_lang} contrasts the two samplers on the eight Gaussian targets.
The deterministic flow reaches a near-optimal distributional approximation to $\rho_\theta^*$ at $k = 50$ and collapses to eight Dirac masses on the critical set of $U_\theta^*$ as $k$ increases.
The predicted stopping time by \eqref{eq:det_stop} during sampling is $k^* = 49$, in agreement with the empirical optimum.
The Langevin sampler reaches the Gibbs invariant measure by $k \approx 100$ and remains on it for the remaining integration steps, with the $k = 200$ panel visually indistinguishable from the $k = 100$ panel.
The figure makes the structural distinction of \eqref{eq:langevin_stop} and \eqref{eq:det_stop} concrete, that is, running the Langevin sampler past its sufficient time is harmless, while running the deterministic flow past $\tau_{\rm det}^*$ strictly degrades the approximation in the manner asserted by Theorem~\ref{thm:no_certificate}.

\section{Composition, Diagnostic OOD, and Outlook}
\label{sec:applications}
 
\subsection{Compositional Generation Under Scalar Parameterization}
\label{sec:compose}
 
Let $U_+,U_- : \R^d\to\R$ be two scalar energy functions trained independently on data sets $\rho_+$ and $\rho_-$. For coefficients $\alpha_+,\alpha_-\in\R$, the additive composition
\begin{equation}
U_{\rm comp}(x) \,:=\, \alpha_+\,U_+(x) \,+\, \alpha_-\,U_-(x)
\label{eq:compose}
\end{equation}
is itself a scalar function on $\R^d$. Therefore, its gradient $\nabla U_{\rm comp} = \alpha_+\,\nabla U_+ + \alpha_-\,\nabla U_-$ is the gradient of a scalar potential, and the closed-loop sampler in (7) driven by drift $-\nabla U_{\rm comp}$ retains an explicit Gibbs invariant measure
\begin{equation}
\rho_{\rm comp}(x) \,\propto\, e^{-\alpha_+\,U_+(x) - \alpha_-\,U_-(x)}.
\label{eq:rho_comp}
\end{equation}
 
\begin{figure}[t]
\centering
\includegraphics[width=\textwidth]{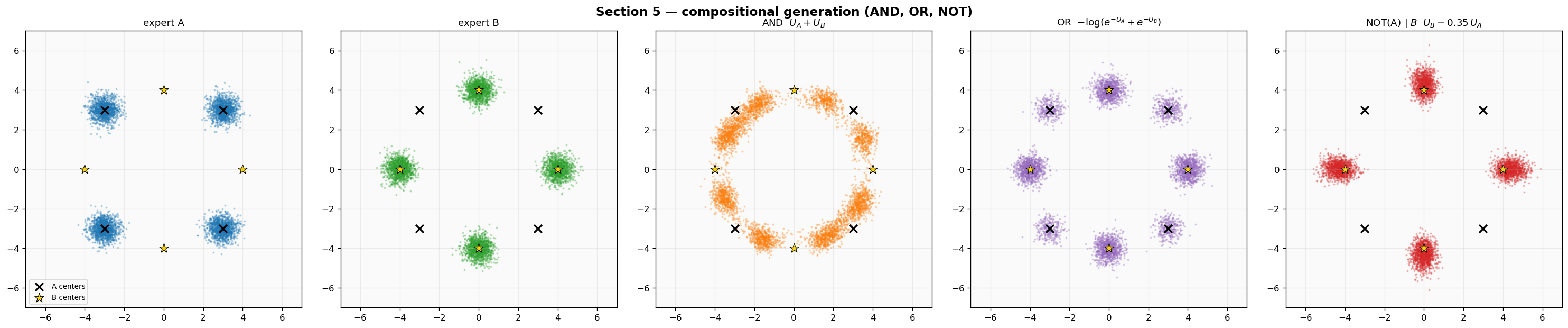}
\caption{Compositional generation by additive energy operations. From left to right: expert $A$ samples, expert $B$ samples, conjunction $U_A + U_B$, disjunction $-\log(e^{-U_A} + e^{-U_B})$, negation $U_B - 0.35\,U_A$. Black crosses mark the $A$ centers, and gold stars mark the $B$ centers. The conjunction places samples on a ring through compromise locations between the two mode sets, since the centers of $A$ and $B$ do not coincide and no single location is a mode of both energies. The negation $U_B - 0.35\,U_A$ produces $4{,}973$ of $5{,}000$ samples within $1.5$ of a $B$ center and $0$ within $1.5$ of any $A$ center, consistent with $A$ modes being repelled by the negative coefficient. Composed energies were not retrained.}
\label{fig:compose}
\end{figure}

 
\paragraph{Empirical verification of compositional generation.}
We train two scalar energies $U_A,U_B$ on disjoint four Gaussian mixtures in $\R^2$ (centers $A = \{(\pm 3,\pm 3)\}$, centers $B = \{(0,\pm 4),(\pm 4,0)\}$) and sample three compositions by Langevin SDE under drift $-\nabla U_{\rm comp}$, with no retraining. The compositions are
\begin{align*}
\textsc{Conjunction}: &\quad U_{\rm comp} \,=\, U_A + U_B,\\
\textsc{Disjunction}: &\quad U_{\rm comp} \,=\, -\log\bigl(e^{-U_A} + e^{-U_B}\bigr),\\
\textsc{Negation~of~A~given~B}: &\quad U_{\rm comp} \,=\, U_B - 0.35\,U_A.
\end{align*}
Figure~\ref{fig:compose} shows $5{,}000$ samples per composition.
The conjunction $U_A + U_B$ does not place samples at a hypothetical intersection of the two mode sets, because no point in $\R^2$ is simultaneously a mode of both energies. Instead, it samples the new minima of $U_A + U_B$, which are compromise locations on a ring at intermediate radius. The disjunction tilts toward $B$ because of a small mismatch in normalization between the two trained energies (alignment offset $0.023$ on a scale of $\approx 62$); the qualitative shape places mass near both mode sets. The negation produces a clean separation: zero samples near $A$ centers, $4{,}973$ of $5{,}000$ near $B$ centers. Across all three compositions, the curl of the composed drift remains at the level of automatic differentiation noise, confirming the structural prediction \eqref{eq:rho_comp}.
 
\subsection{Diagnostic Study: Gradient Norm as an OOD Score}
\label{sec:ood}
 
%
We report a diagnostic study evaluating three scores derived from the closed-loop Langevin dynamics on synthetic and CIFAR~10 data.
We frame this section as a diagnostic study, not as a contribution of new OOD methodology.
The contribution of our study is the systematic evaluation of all three scores on the same scalar EBM.
 
\paragraph{Score hierarchy from the closed-loop.}
Consider three scores
\begin{align*}
    S_1(x) \,=\, U_\theta(x), \qquad
S_2(x) \,=\, \|\nabla U_\theta(x)\|, \qquad
S_3(x) \,=\, \bigl\|x - \Phi^{(\tau)}_{\Delta t}(x)\bigr\|,
\end{align*}
where $\Phi^{(\tau)}_{\Delta t}$ is the $\tau$ step Euler discretization of the deterministic flow $\dot x = -\nabla U_\theta(x)$ with step size $\Delta t$, and $\log Z_\theta$ is computed by thermodynamic integration along the path $U_s = (1-s)U_0 + s U_\theta$ from a Gaussian reference $U_0$. The score $S_1$ is the negative log likelihood up to the constant $\log Z_\theta$. The score $S_2$ is the magnitude of the deterministic descent rate from $\rho_0 = \delta_x$, since $\frac{d}{dt}\E_{\rho_t}U_\theta\big|_{t=0^+} = -\|\nabla U_\theta(x)\|^2$. The score $S_3$ measures trajectory displacement under the deterministic flow, where points near critical regions of $U_\theta$ are not displaced.

\paragraph{CIFAR~10 study.}
We use the checkpoint of \cite{balcerak2025energy}, generation FID $3.34$, with the internal time argument fixed at $t = 0.5$ so that the network acts as a time invariant scalar potential.
The in distribution set is the CIFAR~10 test split with $N = 10{,}000$, while the out of distribution (OOD) sets are SVHN test, the Describable Textures Dataset, and pixel-wise independent uniform noise.
Table~\ref{tab:cifar_ood} reports the OOD AUROC scores.
 
\begin{table}[t]
\centering
\small
\begin{tabular}{lcc}
\toprule
score \,\textbackslash\, OOD set & SVHN & DTD\\
\midrule
$S_1 = V_\theta$ & 0.303 & 0.533\\
$S_2 = \|\nabla V_\theta\|$ & \textbf{0.885} & \textbf{0.641}\\
$S_3$ flow displacement ($\tau=30$) & 0.218 & 0.301\\
\midrule
PixelCNN++ (Yoon et al.\ 2023) & 0.32 & 0.33\\
GLOW (Yoon et al.\ 2023) & 0.24 & 0.27\\
IGEBM \cite{du2019implicit} & 0.63 & 0.48\\
EqM dot product \cite{wang2025eqm} & 0.55 & 0.49\\
\bottomrule
\end{tabular}
\caption{CIFAR~10 OOD results: AUROC for each score on each OOD set. Bold marks the best LEM derived score per column. The score $S_2$ on natural OOD sets (SVHN, DTD) attains $0.763$ on average, exceeding all reported baselines, including the EqM dot product variant by $0.243$ absolute, without retraining or auxiliary heads.}
\label{tab:cifar_ood}
\end{table}

\subsection{Outlook: Control Barrier Functionals on $\PP_2(\R^d)$}
\label{sec:cbf}
 
The control-theoretic perspective taken in this paper opens a path toward formal safety guarantees for generative models in the form of control barrier functionals (CBF) on the Wasserstein space $\PP_2(\R^d)$. In control of finite-dimensional dynamical systems $\dot x = f(x,u)$, a barrier function $b : \R^d\to\R$ encodes a safe set $C = \{x : b(x)\geq 0\}$ and the controller is required to satisfy the pointwise constraint $\nabla b(x)\cdot u(x) + \alpha(b(x)) \geq 0$ for some class $\mathcal{K}$ function $\alpha$, ensuring forward invariance of $C$ along the controlled trajectory \cite{ames2017control}.
 
In our LEM setting, safety is naturally formulated at the distributional level. Given an unsafe set $\mathcal{U}\subset\R^d$ and a tolerance $\beta\in[0,1)$, the candidate barrier functional is
\[
B(\rho) \,=\, \beta - \rho(\mathcal{U}), \qquad \mathcal{S} \,=\, \{\rho\in\PP_2(\R^d) : B(\rho)\geq 0\},
\]
and the lifted CBF condition requires the controlled flow $u^*$ to keep $\rho_t$ within $\mathcal{S}$ for all $t\geq 0$. The conservative structure of scalar parameterization (Section~\ref{sec:compose}) is well aligned with this formulation. That is, a safety penalty implemented by additive energy composition $U_{\rm safe} = U_+ - \lambda\,U_-$ with $U_-$ trained on negative examples retains an explicit Gibbs invariant measure $\rho_{\rm safe}\propto e^{-U_{\rm safe}}$ and inherits the closed-loop guarantees of Section~3.
 
\section{Conclusion}
\label{sec:conclusion}
We have presented a control-theoretic reformulation of static scalar energy generative modeling on the Wasserstein space $\mathcal{P}_2(\mathbb{R}^d)$.
Training and sampling are unified as two instances of one autonomous dynamical dynamics.
The smoothing operation $\hat\rho_{\mathrm{data}} \mapsto \rho^* = \hat\rho_{\mathrm{data}} * K_h$ enters the framework as the prescription for the training target.
Within this single dynamics, two analytic results follow naturally.
The corresponding deterministic gradient flow on the same learned energy landscape admits no distributional Lyapunov certificate, which formalizes the long observed mode collapse phenomenon at the level of the Wasserstein space rather than at the level of individual trajectories.
The stochastic Langevin sampler is able to recover the target distribution under the same Lyapunov function, and the framework supplies stopping criteria for both samplers that distinguish a sufficient budget from a hard cutoff.
The deeper insight is that mode collapse and stochastic correctness are not two separate phenomena, but two consequences of whether the noise is present in the controller.

The structural payoff of the reformulation is that the static scalar energy paradigm now sits inside the same conceptual setting as nonlinear control theory.
Scalar energies compose additively while preserving the Gibbs invariant form and the closed-loop Lyapunov certificate, so the additive composition operation needed to combine trained energies is algebraically free of structural cost.
This is the gateway through which control barrier functionals on the Wasserstein space encode safety as forward invariance of a sublevel set, and contraction metrics, nonlinear observers, and stochastic safety certificates become accessible to generative modeling.
We view the present paper as the foundational layer that makes such extensions possible for the static scalar energy paradigm, and as an invitation to read generative modeling as a chapter of nonlinear control theory rather than a discipline parallel to it.

\bibliographystyle{plainnat}
\bibliography{ref}

\appendix

\section{Proof Details}
\label{app:proofs}

\subsection{Lyapunov Derivative Computation}
\label{app:lyapunov_deriv}

Since $\rho^* = \exp(-U^*)/Z$, we have $\log \rho^*(x) = -U^* - \log Z$ and thus $\log(\rho_\theta/\rho^*) = \log \rho_\theta + U^* + \log Z$.
Differentiating $V(\rho_\theta) = \int _{\R^d} \rho_\theta(\log \rho_\theta + U^* + \log Z)\,dx$ with respect to $t$ yields
\begin{equation}
  \frac{d}{dt}V(\rho_\theta) = \frac{d}{dt}\int_{\R^d} \rho_\theta \log\rho_\theta dx +\frac{d}{dt}\int_{\R^d} \rho_\theta U^* dx + \frac{d}{dt}\int_{\R^d} \rho_\theta \log Z dx.
\end{equation}
The first term gives
\begin{equation}
 \frac{d}{dt}\int_{\R^d} \rho_\theta \log \rho_\theta dx = \int_{\R^d} (\partial_t \rho_t)\log \rho_\theta dx + \int_{\R^d} \partial_t \rho_\theta dx,
\end{equation}
where $\int \partial_t \rho_\theta dx =0$ by conservation of mass $\int \rho_\theta dx = 1$.
Since $U^*$ is time independent, the second term gives $\frac{d}{dt}\int \rho_\theta U^* dx = \int (\partial_t \rho_\theta) U^* dx$.
Since $\log Z$ is constant, combining these results gives
\begin{equation}
  \frac{d}{dt}V(\rho_\theta)=\int_{\R^d} (\partial_t \rho_\theta)\bigl(\log \rho_\theta + U^*\bigr) dx.
\end{equation}
Using the continuity equation $\partial_t \rho_\theta = -\nabla \cdot (\rho_\theta u)$ and integration by parts with parts $ \rho_\theta u$ and $ \log \rho_\theta + U^*$ (boundary terms vanish with the decay of $\rho_t$ at infinity) yields 
\begin{align}
  \frac{d}{dt}V(\rho_\theta) &= -\int_{\R^d} \nabla \cdot (\rho_\theta u)  \bigl(\log \rho_\theta + U^* \bigr)\,dx \nonumber\\
  &= 
  \int_{\R^d} \rho_\theta u \cdot \nabla\bigl(\log \rho_\theta + U^* \bigr)\,dx = \int_{\R^d} \rho_\theta u \cdot \bigl(\nabla \log \rho_\theta + \nabla U^*\bigr)\,dx.
\end{align}

\subsection{Fixed Point Analysis}
\label{app:fixed_point}

{For learned results $\rho^*_\theta, U^*_\theta$, and $Z^*_\theta$, we remove their scripts below for simplicity.} Substituting the Gibbs density $\rho = \frac{e^{-U}}{Z}$ into the deterministic flow term $\nabla \cdot (\rho \nabla U)$ yields
\begin{equation}
  \nabla \cdot (\rho \nabla U) \;=\; \nabla \rho \cdot \nabla U + \rho\,\Delta U.
\end{equation}
Using the chain rule, we have
\begin{equation}
  \nabla \rho  = \nabla \bigl(\frac{e^{-U}}{Z} \bigr) = - \bigl(\frac{e^{-U}}{Z} \bigr) \nabla U =- \rho \nabla U.
\end{equation}
Then we write
\begin{equation}
  \nabla \cdot (\rho \nabla U) \;=\; -\rho\,\|\nabla U\|^2 + \rho\,\Delta U \;=\; \rho\bigl(\Delta U - \|\nabla U\|^2\bigr).
\end{equation}

\subsection{KL Divergence Decay Along Deterministic Flow}
\label{app:stein}

With the deterministic flow in \eqref{eq:det_flow} and the corresponding controller $u^*_d(x)= -\nabla U_\theta^* (x)$, the Lyapunov derivative in \eqref{eq:lie_derivative} becomes
\begin{align}\label{a4.1}
  \frac{d \KL(\rho_t\| \rho^*_\theta)}{dt} &= 
  -\int_{\R^d} \rho_t\, \nabla U^*_\theta \cdot \bigl(\nabla \log \rho_t + \nabla U^*_\theta \bigr)\,dx\nonumber\\
  &=-\int_{\R^d} \rho_t\, \nabla U^*_\theta \cdot \nabla \log \rho_t \ dx -\int_{\R^d} \rho_t\, \| \nabla U^*_\theta \|^2\,dx.
\end{align}
Since $\rho_t \nabla \log \rho_t = \nabla \rho_t$, the first term becomes
\begin{align} 
  &-\int_{\R^d} \rho_t\, \nabla U^*_\theta \cdot \nabla \log \rho_t \ dx = -\int_{\R^d} \nabla U^*_\theta \cdot \nabla \rho_t \ dx = \int_{\R^d} \rho_t \Delta U^*_\theta \ dx.
\end{align}
Using this result in \eqref{a4.1} yields
\begin{align}
  \frac{d \KL(\rho_t\| \rho^*_\theta)}{dt} &= \int_{\R^d} \rho_t \Delta U^*_\theta \ dx
   -\int_{\R^d} \rho_t\, \| \nabla U^*_\theta \|^2\,dx,
\end{align}
which is equivalent to \eqref{eq:det_lyap}.

\subsection{Proof of Theorem~\ref{thm:no_certificate}}
\label{app:det_proofs}

Suppose there exists a Lyapunov functional $V : \PP_2(\R^d) \to \R_{\ge 0}$ satisfying conditions (i)-(iii).
Consider the deterministic flow initialized at the target density, i.e., $\rho_0 = \rho^*_\theta$.
By condition (ii), $V(\rho_t) \geq 0$ for all $\rho \in \PP_2(\R^d)$, and by condition (i), $ V(\rho^*_\theta) = 0$. Condition (iii) implies that $V(\rho_t) = 0$ is nonincreasing along trajectories of the deterministic flow. Therefore,
\begin{equation}
  V(\rho_t) \leq V(\rho_\theta^*) = 0, \ \text{for all }t \geq 0.
\end{equation}
Since $V\geq 0$, $V(\rho_t)  = 0$ for all $t \geq 0$. By condition (i), this implies $\rho_t = \rho_\theta^*$ for all $t \ge 0$, so the target density $\rho_\theta^*$ must be a stationary solution (a fixed point) of the deterministic flow.
From the continuity equation in \eqref{eq:det_flow}, stationarity of $\rho^*_\theta$ requires $\nabla \cdot (\rho_\theta^* \nabla U^*_\theta) =0$. By Appendix \ref{app:fixed_point}, this condition is equivalent to $h(x) = \Delta U^*_\theta(x) - \|\nabla U^*_\theta(x)\|^2 = 0$ for all $x$.

We now show that $h(x)$ cannot vanish identically.
Let $x_0$ be any critical point such that $\nabla U^*_\theta(x_0) = 0$. Then, $h(x_0) = \Delta U^*_\theta(x_0) = \tr(\nabla^2 U^*_\theta(x_0))$.
At any nondegenerate local minimum, the Hessian $\nabla^2 U^*_\theta(x_0)$ is positive definite, and therefore 
\begin{equation}
  h(x_0) = \tr(\nabla^2 U^*_\theta(x_0)) > 0.
\end{equation}
In addition, by the assumption in Theorem \ref{thm:no_certificate}, i.e., $\|\nabla U^*_\theta(x)\|^2 \to \infty$ as $\|x\| \to \infty$, the Lipschitz gradient assumption implies that $\|\nabla U^*_\theta(x)\|^2$ is bounded, and hence $|\Delta U_\theta^*(x)| = |\tr(\nabla^2 U_\theta^*(x))| \le dL$ for some constant $L>0$. Therefore,
\begin{equation}
  h(x) = \Delta U_\theta^*(x) - \|\nabla U_\theta^*(x)\|^2 \to -\infty, \ \text{as} \ \|x\| \to \infty
\end{equation}
Since $h$ is continuous, positive at any nondegenerate minimum, and tends to $-\infty$ at infinity, $h$ must change sign and consequently is not identically zero, contradicting the stationarity requirement.


\end{document}